\documentclass[11pt,letterpaper]{article}
\usepackage{acl2002}
\usepackage{times}
\usepackage{latexsym}
\usepackage{amsmath}
\usepackage{epsfig}
\title{Unsupervised Discovery of Morphemes}
\author{{\bf Mathias Creutz} \and {\bf Krista Lagus}\\
Neural Networks Research Centre\\
Helsinki University of Technology\\
P.O.Box 9800, FIN-02015 HUT, Finland\\ 
{\tt \{Mathias.Creutz, Krista.Lagus\}@hut.fi}
}

\date{}

\begin{document}
\maketitle
\begin{abstract}
  We present two methods for unsupervised segmentation of
  words into morpheme-like units. The model utilized is especially suited
  for languages with a rich morphology, such as Finnish.
  The first method is based on the Minimum Description Length (MDL)
  principle and works online.
  In the second method, Maximum Likelihood (ML) optimization is used.
  The quality of the segmentations is measured using an
  evaluation method that compares the segmentations produced to an
  existing morphological analysis.
  Experiments on both Finnish and English
  corpora show that the presented methods perform 
  well compared to a current state-of-the-art system.
\end{abstract}

\section{Introduction}
\label{sec:intro}

According to linguistic theory, morphemes are considered to be the
smallest meaning-bearing elements of language, and they can be defined
in a language-independent manner. However, no adequate
language-inde\-pendent definition of the \emph{word} as a unit has
been agreed upon \nocite{Karlsson98} (Karlsson, 1998, p.~83). If
effective methods can be devised for the unsupervised discovery of
morphemes, they could aid the formulation of a linguistic theory of
morphology for a new language.

It seems that even approximative automated morphological analysis 
would be beneficial for many
natural language applications dealing with large vocabularies. For
example, in text retrieval it is customary to preprocess
texts by returning words to their base forms, especially for
morphologically rich languages. 


Moreover, in large vocabulary speech recognition, predictive models of
language are typically used for selecting the most plausible words
suggested by an acoustic speech recognizer (see, e.g., Bellegarda, 
2000\nocite{bellegarda00.ieee}).
Consider, for example the estimation of the standard $n$-gram model,
which entails the estimation of the probabilities of all sequences of
$n$ words.  When the vocabulary is very large, say 100~000 words, 
the basic problems in the estimation of the language model
are: (1) If words are used as basic representational units in the
language model, the number of basic units is very high and the
estimated word $n$-grams are poor due to sparse data. (2) Due to the
high number of possible word forms, many perfectly valid word forms
will not be observed at all in the training data, even in large
amounts of text. These problems are particularly severe for languages
with rich morphology, such as Finnish and Turkish. For example, in
Finnish, a single verb may appear in thousands of different
forms \cite{Karlsson87}.

The utilization of morphemes as basic representational units in a
statistical language model instead of words seems a promising course. Even a
rough morphological segmentation could then be sufficient.
On the other hand, the construction of a comprehensive morphological analyzer
for a language based on linguistic theory requires 
a considerable amount of work by experts. 
This is both slow and expensive and therefore not applicable to all
languages. The problem is further compounded as languages evolve, new 
words appear and grammatical changes take place.
Consequently, it is important to develop methods that are able to
\emph{discover} a morphology for a language based on unsupervised
analysis of large amounts of data.

As the morphology discovery from untagged corpora is a computationally
hard problem, in practice one must make some assumptions about the
structure of words. 
%
The appropriate specific assumptions are somewhat
language-dedependent. For example, for English it may be useful to
assume that words consist of a stem, often followed by a suffix and
possibly preceded by a prefix. By contrast, a Finnish word typically
consists of a stem followed by multiple suffixes. In addition,
compound words are common, containing an alternation of stems and
suffixes, e.g., the word \texttt{kahvinjuojallekin} (Engl. '\emph{also for
[the] coffee drinker}'; cf. Table \ref{table:morphology})\footnote{For
a comprehensive view of Finnish morphology, see \cite{Karlsson87}.}.
Moreover, one may ask, whether a morphologically complex word exhibits
some hierarchical structure, or whether it is merely a flat
concatenation of stems and suffices.

\begin{table}
\caption{The morphological structure of the Finnish word for `\emph{also for
[the] coffee drinker}'.}
\label{table:morphology}
\vskip 0.12in
\begin{tabular}{|l|l|l|l|l|l|l|}
\hline
Word & \multicolumn{6}{|l|}{kahvinjuojallekin} \\
\hline
Morphs & kahvi & n & juo & ja & lle & kin \\
\hline
Transl. & coffee & of & drink & -er & for & also \\
\hline
\end{tabular}
\end{table}

\subsection{Previous Work on Unsupervised Segmentation}
Many existing morphology discovery algorithms concentrate on
identifying prefixes, suffixes and stems, i.e., assume a rather
simple inflectional morphology.

D\'ejean (1998) \nocite{deJean98} concentrates on the problem of
finding the list
of frequent affixes for a language rather than attempting to produce a
morphological analysis of each word. Following the work of Zellig
Harris he identifies possible morpheme boundaries by looking at the
number of possible letters following a given sequence of letters, and
then utilizes frequency limits for accepting morphemes.

Goldsmith (2000) \nocite{Goldsmith00} concentrates on stem+suffix-languages,
in particular Indo-European languages, and tries to produce output
that would match as closely as possible with the analysis given by a
human morphologist. He further assumes that stems form groups that he calls
\emph{signatures}, and each signature shares a set of possible affixes. 
He applies an MDL criterion for model optimization.

The previously discussed approaches consider only individual words
without regard to their contexts, or to their semantic content. In a
different approach, Schone and Jurafsky (2000) \nocite{SchoneJurafsky00} utilize the context of each term
to obtain a semantic representation for it using LSA. The division to
morphemes is then accepted only when the stem and stem+affix
are sufficiently similar semantically. Their method is shown to
improve on the performance of Goldsmith's \emph{Linguistica} on CELEX,
a morphologically analyzed English corpus.

In the related field of text segmentation, one can sometimes
obtain morphemes. Some of the approaches remove spaces from text and
try to identify word boundaries utilizing e.g. entropy-based measures,
as in \cite{Redlich93}. 

Word induction from natural language text without word boundaries is
also studied in \cite{DeligneBimbot97,Hua00}, where MDL-based model
optimization measures are used. Viterbi or the forward-backward
algorithm
(an EM algorithm) is used for improving the segmentation of the
corpus\footnote{The regular EM procedure only maximizes the
likelihood of the data. To follow the MDL approach where model cost is
also optimized, Hua includes the model cost as a penalty term on pure
ML probabilities.}.


Also de Marcken (1995; 1996) \nocite{deMarcken95,deMarcken96} studies
the problem of learning a lexicon, but instead of optimizing the cost
of the whole corpus, as in \cite{Redlich93,Hua00}, de Marcken starts
with sentences. Spaces are included as any other characters.

Utterances are also analyzed in \cite{Kit99} where optimal
segmentation for an utterance is sought so that the compression effect
over the segments is maximal. The compression effect is measured in
what the authors call Description Length Gain, defined as the relative
reduction in entropy. The Viterbi algorithm is used for searching for
the optimal segmentation given a model.  The input utterances include
spaces and punctuation as ordinary characters. The method is evaluated
in terms of precision and recall on word boundary prediction.

Brent presents a general, modular probabilistic model structure for
word discovery \cite{Brent99}. He uses a minimum representation length
criterion for model optimization and applies an incremental, greedy
search algorithm which is suitable for on-line learning such that
children might employ.


\subsection{Our Approach}

In this work, we use a model where words may consist of lengthy
sequences of segments.  This model is especially suitable for
languages with agglutinative morphological structure.  We call the
segments \emph{morphs} and at this point no distinction is made
between stems and affixes.

The practical purpose of the segmentation is to provide a vocabulary
of language units that is smaller and generalizes better than a
vocabulary consisting of words as they appear in text.  Such a
vocabulary could be utilized in statistical language modeling,
e.g., for speech recognition.  Moreover, one could assume that such a
discovered morph vocabulary would correspond rather closely to 
linguistic morphemes of the language.

We examine two methods for unsupervised learning of the model,
presented in Sections~\ref{sec:method1} and~\ref{sec:method2}.
The cost function for the first method is derived from the Minimum Description
Length principle from classic information theory \cite{Rissanen89},
which simultaneously measures the goodness of the representation and
the model complexity.  Including a model
complexity term generally improves generalization by inhibiting
overlearning, a problem especially severe for sparse data.  An
incremental (online) search algorithm is utilized that applies a
hierarchical splitting strategy for words.
In the second method the cost
function is defined as the maximum likelihood of the data given the
model.  Sequential splitting is applied and a batch learning
algorithm is utilized.

In Section~\ref{sec:eval}, we develop a method for evaluating the
quality of the morph segmentations produced by the unsupervised
segmentation methods. Even though the morph segmentations obtained are
not intended to correspond exactly to the morphemes of linguistic
theory, a basis for comparison is provided by existing, linguistically
motivated morphological analyses of the words.

Both segmentation methods are applied to the segmentation of both
Finnish and English words. In Section~\ref{sec:experiments}, we
compare the results obtained from our methods to results produced by
Goldsmith's \emph{Linguistica} on the same data. 

\section{Method 1: Recursive Segmentation and MDL Cost}
\label{sec:method1}
%

The task is to find the optimal segmentation of the source text into
morphs.  One can think of this as constructing a model of the data in
which the model consists of a vocabulary of morphs, i.e. the
\emph{codebook} and the data is the sequence of text.  We try to find
a set of morphs that is concise, and moreover gives a concise
representation for the data. This is achieved by utilizing an MDL cost
function.
\subsection{Model Cost Using MDL}
The total cost consists of two parts: the cost of the source text in
this model and the cost of the codebook.  
Let $M$ be the morph codebook (the vocabulary of morph types) and $D =
m_1m_2\ldots m_n$ the sequence of morph tokens that makes up the
string of words. We then define the total cost $C$ as
\begin{eqnarray*}
C &=&\text{Cost}(\text{Source text}) + \text{Cost}(\text{Codebook}) \\
& = & \sum_{\text{tokens}} -\log p(m_i) + 
 \sum_{\text{types}} k * l(m_j) 
\label{eqn:MDLcost}
\end{eqnarray*}
The cost of the source text is thus the negative log\-likelihood of the
morph, summed over all the morph tokens that comprise the source text.
The cost of the codebook is simply the length in bits needed to represent
each morph separately as a string of characters, summed over the
morphs in the codebook.  The length in characters of
the morph $m_j$ is denoted by $l(m_j)$ and $k$ is the number of bits
needed to code a character
(we have used a value of 5 since that is sufficient for coding 32
lower-case letters).  For $p(m_i)$ we use the ML estimate, i.e., the
token count of $m_i$ divided by the total count of morph tokens.

\subsection{Search Algorithm}

The online search algorithm works by incrementally suggesting 
changes that could 
improve the cost function.  Each time a new word token is read
from the input, different ways of segmenting it into morphs are
evaluated, and the one with minimum cost is selected. 

\paragraph{Recursive segmentation.} 
The search for the optimal morph segmentation proceeds recursively.
First, the word as a whole is considered to be a morph and added to the
codebook. Next, every possible split of the word into two parts is
evaluated.  

The algorithm selects the split (or no split) that yields the minimum
total cost. In case of no split, the processing of the word is
finished and the next word is read from input. Otherwise, the search
for a split is performed recursively on the two segments. The order of splits can be represented
as a binary tree for each word, where the leafs represent the morphs
making up the word, and the tree structure describes the ordering of the
splits.

\begin{figure}[ht]
\centerline{\includegraphics[width=0.44\textwidth]{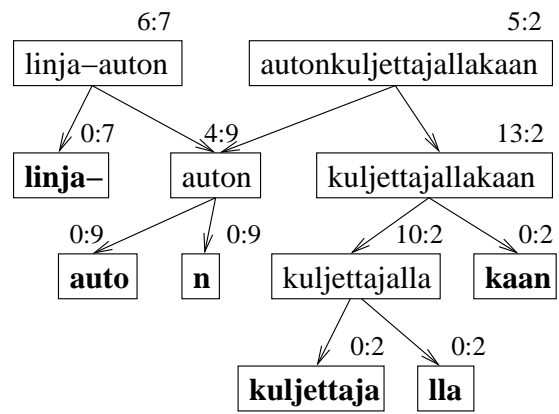}}
\caption{Hierarchical structure of the segmentation of the words
\texttt{linja-auton} and \texttt{autonkuljettajallakaan}. The boxes
represent chunks. Boxes with bold text are morphs, and are
part of the codebook. The numbers above each box are the split location
(to the left of the colon sign) and the
occurrence count of the chunk (to the right of the colon sign).}
\label{fig:Autonkuljettaja}
\end{figure}

During model search, an overall hierarchical data structure is used for keeping
track of the current segmentation of every word type encountered so
far. Let us assume that we have seen seven instances of
\texttt{linja-auton} (Engl.~'\textit{of [the] bus}') and two instances
of \texttt{auton\-kuljetta\-jalla\-kaan} (Engl.~'\textit{not even
by/at/with [the] car driver}').  Figure \ref{fig:Autonkuljettaja}
then shows a possible structure used for representing the
segmentations of the data.  Each chunk is provided with an occurrence
count of the chunk in the data set and the split location in this
chunk. A zero split location denotes a leaf node, i.e., a morph.
The occurrence counts flow down through the
hierachical structure, so that the count of a child always
equals the sum of the counts of its parents. The occurrence
counts of the leaf nodes are used for computing the relative
frequencies of the morphs.
To find out the morph sequence that a
word consists of, we look up the chunk that is identical to the word,
and trace the split indices recursively until we reach the leafs,
which are the morphs. 

Note that the hierarchical structure is used only during model search:
It is not part of the final model, and accordingly no cost is associated 
with any other nodes than the leaf nodes.

\paragraph{Adding and removing morphs.}
Adding new morphs to the codebook increases the codebook cost.
Consequently, a new word token will tend to be split
into morphs already listed in the codebook, which may lead to
local optima.  To better escape local optima, each time a new word
token is encountered, it is resegmented, whether or not this word has been
observed before. If the word has been observed (i.e. the corresponding
chunk is found in the hierarchical structure), we first \emph{remove}
the chunk and decrease the counts of all its children.  Chunks with
zero count are removed (remember that removal of leaf nodes
corresponds to removal of morphs from the codebook).  Next, we
increase the count of the observed word chunk by one and re-insert it
as an unsplit chunk. Finally, we apply the recursive splitting to the
chunk, which may lead to a new, different segmentation of the word.

\begin{figure}[htb]
\includegraphics[width=0.47\textwidth]{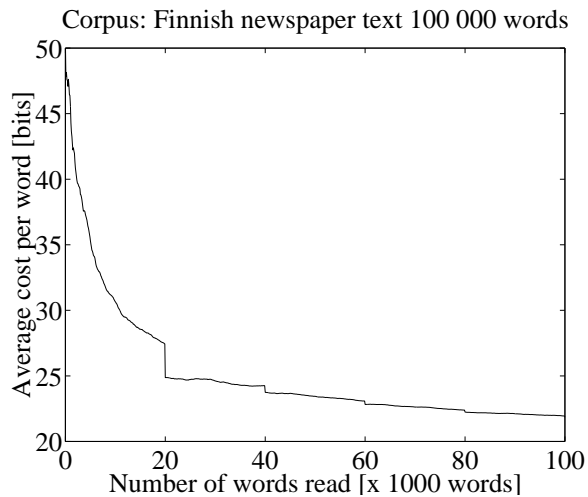}
\caption{Development of the average word cost when processing
newspaper text. Dreaming, i.e., the  re-processing of the words
encountered so far, takes place five times, which can be seen as
sudden drops on the curve.}  
\label{fig:CostDevelopment}
\end{figure}

\paragraph{``Dreaming''.}
Due to the online learning, as the number of processed words
increases, the quality of the set of morphs in the codebook gradually
improves. Consequently, words encountered in the beginning of the
input data, and not observed since, may have a sub-optimal
segmentation in the new model, since at some point more suitable
morphs have emerged in the codebook.  We have therefore introduced a
'dreaming' stage: At regular intervals the system stops reading words
from the input, and instead iterates over the words already
encountered in random order. These words are resegmented and 
thus compressed further,
if possible. Dreaming continues for a limited time or until no 
considerable decrease in the total cost can be observed.
Figure \ref{fig:CostDevelopment} shows the development of the average
cost per word as a function of the increasing amount of source text.

\section{Method 2: Sequential Segmentation and ML Cost}
\label{sec:method2}
\subsection{Model Cost Using ML}
In this case, we use as cost function the likelihood of the data,
i.e., $P(data|model)$. Thus, the model cost is not included.  This
corresponds to Maximum-Likelihood (ML) learning. The cost is then
\begin{equation}
\text{Cost}(\text{Source text}) = \sum_{\text{morph tokens}}
-\log p(m_i),
\end{equation}
where the summation is over all morph tokens in the source data.  
As before, for
$p(m_i)$ we use the ML estimate, i.e., the token count of $m_i$
divided by the total count of morph tokens.

\subsection{Search Algorithm}

In this case, we utilize batch learning where an EM-like
(Expectation-Maximization) algorithm is used for optimizing the model.
Moreover, splitting is not recursive but proceeds linearly.

\begin{enumerate}
  \item Initialize segmentation by splitting words into morphs at
  random intervals, starting from the beginning of the word.  The
  lengths of intervals are sampled from the Poisson distribution with
  $\lambda = 5.5$. If the interval is larger than the number of
  letters in the remaining word segment, the splitting ends.

  \item Repeat for a number of iterations:
	\begin{enumerate}
	  \item Estimate morph probabilities for the given splitting. 

	  \item Given the current set of morphs and their probabilities, 
		re-segment the text using the Viterbi algorithm for finding the
		segmentation with lowest cost for each word.
          \item If not the last iteration: Evaluate the segmentation
		of a word against rejection criteria. 
		If the proposed segmentation is not accepted, segment this
		word randomly (as in the Initialization step).  
\end{enumerate} \end{enumerate}
Note that the possibility of introducing a random segmentation at step
$(c)$ is the only thing that allows for the addition of new morphs. (In the
cost function their cost would be infinite, due to ML probability
estimates). In fact, without this step the algorithm seems to get
seriously stuck in suboptimal solutions.

\paragraph{Rejection criteria.}
(1) Rare morphs. Reject the segmentation of a word if the segmentation
contains a morph that was used in only one word type in the previous iteration.
This is motivated by the fact that extremely rare morphs are often
incorrect.
(2) Sequences of one-letter morphs.  Reject the segmentation if it
contains two or more one-letter morphs in a sequence. For instance,
accept the segmentation \texttt{halua + n} (Engl. '\textit{I want}',
i.e. present stem of the verb '\textit{to want}' followed by the
ending for the first person singular), but reject the segmentation
\texttt{halu + a + n} (stem of the noun '\textit{desire}' followed by a
strange sequence of endings). Long sequences of one-letter morphs
are usually a sign of a very bad local optimum that may even get worse
in future iterations, in case too much probability mass is transferred
onto these short morphs\footnote{Nevertheless, for Finnish there do
exist some one-letter morphemes that can occur in a sequence. However,
these morphemes can be thought of as a group that belongs together:
e.g., the Finnish \texttt{talo + j + a} (plural partitive of 'house');
can also be thought of as \texttt{talo + ja}.}.



\section{Evaluation Measures}
\label{sec:eval}
We wish to evaluate the method quantitatively from the following perspectives: (1)
correspondence with linguistic morphemes, (2) efficiency of
compression of the data, and (3) computational efficiency.
The efficiency of compression can be evaluated as the total
description length of the corpus and the codebook (the MDL cost
function).  The computational efficiency of the algorithm can be
estimated from the running time and memory consumption of the program.
However, the linguistic evaluation is in general not so straightforward.

\subsection{Linguistic Evaluation Procedure}
If a corpus with marked morpheme boundaries is available, the
linguistic evaluation can be computed as the precision and recall of
the segmentation. Unfortunately, we did not have such data
sets at our disposal, and for Finnish such do not even
exist. In addition, it is not always clear exactly where the
morpheme boundary should be placed. Several alternatives may be
possible, cf. Engl. \texttt{hope + d} vs. \texttt{hop + ed}, (past tense of
\emph{to hope}).  

Instead, we utilized an existing tool for providing a morphological
analysis, although not a segmentation, of words, based on the
two-level morphology of Koskenniemi (1983)\nocite{Koskenniemi83}. The analyzer is a
finite-state transducer that reads a word form as input and outputs
the base form of the word together with grammatical tags. Sample
analyses are shown in Figure~\ref{fig:TWOLAnalyses}.

\begin{figure}[htb]
\begin{tabular}{|l|l|l|}
\hline
Input & \multicolumn{2}{|l|}{Output} \\
\hline
Word & Base form & Tags \\
\hline
\hline
easily & \textsc{easy} & \texttt{$<$DER:ly$>$ ADV} \\
\hline
bigger & \textsc{big} & \texttt{A CMP} \\
\hline
hours' & \textsc{hour} & \texttt{N PL GEN} \\
\hline
\hline
auton & \textsc{auto} & \texttt{N SG GEN} \\
\hline
puutaloja & \textsc{puu\#talo} & \texttt{N PL PTV} \\
\hline
tehnyt & \textsc{tehd\"a} & \texttt{V ACT PCP2 SG} \\
\hline
\end{tabular}
\caption{Morphological analyses for some English and Finnish word
forms. The Finnish words are \texttt{auton} (\textit{car's}),
\texttt{puutaloja} (\textit{[some] wooden houses}) and \texttt{tehnyt}
(\textit{[has] done}). The tags are \texttt{A} (adjective),
\texttt{ACT} (active voice), \texttt{ADV} (adverb), \texttt{CMP}
(comparative), \texttt{GEN} (genitive), \texttt{N} (noun),
\texttt{PCP2} (2nd participle), \texttt{PL} (plural), \texttt{PTV}
(partitive), \texttt{SG} (singular), \texttt{V} (verb), and
\texttt{$<$DER:ly$>$} (-ly derivative).}
\label{fig:TWOLAnalyses}
\end{figure}

The tag set consists of tags corresponding to morphological affixes
and other tags, for example, part-of-speech tags.  We preprocessed the
analyses by removing other tags than those corresponding to affixes,
and further split compound base forms (marked using the \# character
by the analyzer) into their constituents.  As a result, we obtained for each
word a sequence of labels that corresponds well to a linguistic
\emph{morphemic analysis} of the word.  A label can 
often be considered to correspond to a single word segment, and the
labels appear in the order of the segments.

The following step consists in retrieving the segmentation produced by
one of the unsupervised segmentation algorithms, and trying to align
this segmentation with the desired morphemic label sequence
(cf. Figure~\ref{fig:Alignments}).

A good segmentation algorithm will produce morphs that align gracefully
with the correct morphemic labels, preferably producing a
one-to-one mapping. A one-to-many mapping from morphs to labels is
also acceptable, when a morph forms a common entity,
such as the suffix \textit{-ja} in \textit{puutaloja}, which contains
both the plural and partitive element. By contrast, a many-to-one
mapping from morphs to a label is a sign of excessive
splitting, e.g., \texttt{t + alo} for \textit{talo} (cf. English
\texttt{h + ouse} for \textit{house}). 



\begin{figure}[htb]
\begin{tabular}{|l|l|l|l|l|l|}
\hline
Correct labels & \multicolumn{3}{|l|}{\textsc{big}} & \multicolumn{2}{|l|}{\texttt{CMP}} \\
\hline
Morph sequence & \multicolumn{3}{|l|}{\texttt{bigg}} & \multicolumn{2}{|l|}{\texttt{er}} \\
\hline
\hline
Correct labels & \multicolumn{3}{|l|}{\textsc{hour}} & \texttt{PL} & \texttt{GEN} \\
\hline
Morph sequence & \multicolumn{3}{|l|}{\texttt{hour}} & \texttt{s} & \texttt{'} \\
\hline
\hline
Correct labels & \textsc{puu} & \multicolumn{2}{|l|}{\textsc{talo}} & \texttt{PL} & \texttt{PTV} \\
\hline
Morph sequence & \texttt{puu} & \texttt{t} & \texttt{alo} & \multicolumn{2}{|c|}{\texttt{ja}} \\
\hline
\end{tabular}
\caption{Alignment of obtained morph sequences with their respective
correct morphemic analyses. We assume that the segmentation
algorithm has split the word \textit{bigger} into the morphs
\texttt{bigg + er}, \textit{hours'} into \texttt{hour + s + '} and
\textit{puutaloja} into \texttt{puu + t + alo + ja}.}
\label{fig:Alignments}
\end{figure}

\paragraph{Alignment procedure.}
We align the morph sequence with the morphemic label sequence using
dynamic programming, namely Viterbi alignment, to find the best
sequence of mappings between morphs and morphemic labels.  Each
possible pair of morph/morphemic label has a distance associated with
it. For each segmented word, the algorithm searches for the alignment
that minimizes the total alignment distance for the word. The distance
$d(M, L)$ for a pair of morph $M$ and label $L$ is given by:
\begin{equation}
d(M, L) = -\log \frac{c_{M, L}}{c_{M}},
\end{equation}
where $c_{M,L}$ is the number of word tokens in which the morph $M$
has been aligned with the label $L$; and $c_{M}$ is the number of word
tokens that contain the morph $M$ in their segmentation. The distance
measure can be thought of as the negative logarithm of a conditional
probability $P(L|M)$. This indicates the probability that a morph
$M$ is a realisation of a morpheme represented by the label $L$. Put
another way, if the unsupervised segmentation algorithm discovers
morphs that are allomorphs of real morphemes, a particular allomorph
will ideally always be aligned with the same (correct) morphemic
label, which leads to a high probability $P(L|M)$, and a short
distance $d(M, L)$\footnote{This holds especially for allomorphs of
'stem morphemes', e.g., it is possible to identify the English
morpheme \textit{easy} with a probability of one from both its
allomorphs: \textit{easy} and \textit{easi}. However, suffixes, in
particular, can have several meanings, e.g., the English suffix
\textit{s} can mean either the plural of nouns or the third person
singular of the present tense of verbs.}. In contrast, if the
segmentation algorithm does not discover meaningful morphs, each of
the segments will be aligned with a number of different morphemic
labels throughout the corpus, and as a consequence, the probabilities
will be low and the distances high.

We then utilize the EM algorithm for iteratively improving the alignment.
The initial alignment that is used for computing initial distance values
is obtained through a string matching procedure: String matching is
efficient for aligning the stem of the word with the base form (e.g.,
the morph \texttt{puu} with the label \textsc{puu}, and the morphs
\texttt{t + alo} with the label \textsc{talo}). The suffix morphs that
do not match well with the base form labels will end up aligned somehow with
the morphological tags (e.g., the morph \texttt{ja} with the labels
\texttt{PL + PTV}).

\paragraph{Comparison of methods.}
In order to compare two segmentation algorithms, the segmentation of
each is aligned with the linguistic morpheme labels, and the total
distance of the alignment is computed. Shorter total distance indicates
better segmentation.

However, one should note that the distance measure used favors long
morphs. If a particular ``segmentation'' algorithm does not split one
single word of the corpus, the total distance can be zero. In such a
situation, the single morph that a word is composed of is aligned with
all morphemic labels of the word. The morph $M$, i.e., the word, is
unique, which means that all probabilities $P(L|M)$ are equal to one:
e.g., the morph \texttt{puutaloja} is always aligned with the labels
\textsc{puu} + \textsc{talo} + \texttt{PL} + \texttt{PTV} and no other
labels, which yields the probabilities $P(\mathtt{PUU} \mid
\mathtt{puutaloja}) = P(\mathtt{TALO} \mid \mathtt{puutaloja}) =
P(\mathtt{PL} \mid \mathtt{puutaloja}) = P(\mathtt{PTV} \mid
\mathtt{puutaloja}) = 1$.

Therefore, part of the corpus should be used as training data, and
the rest as test data. Both data sets are segmented using the
unsupervised segmentation algorithms. The training set is then used
for estimating the distance values $d(M, L)$. These values are used
when the test set is aligned. The better segmentation algorithm is the
one that yields a better alignment distance for the test set.

For morph/label pairs that were never observed in the training
set, a maximum distance value is assigned. A
good segmentation algorithm will find segments that are good building
blocks of entirely new word forms, and thus the maximum distance
values will occur only rarely.

\section{Experiments and Results}
\label{sec:experiments}

We compared the two proposed methods as well as Goldsmith's program
\emph{Linguistica}\footnote{http://humanities.uchicago.edu/faculty/goldsmith/Linguist\-ica2000/} on both Finnish and English corpora.  The Finnish
corpus consisted of newspaper text from
CSC\footnote{http://www.csc.fi/kielipankki/}. A morphosyntactic
analysis of the text was performed using the Conexor FDG
parser\footnote{http://www.conexor.fi/}. All characters were converted
to lower case, and words containing other characters than \emph{a}
through \emph{z} and the Scandinavian letters \emph{\aa}, \emph{\"a}
and \emph{\"o} were removed. Other than morphemic tags were removed
from the morphological analyses of the words. The remaining tags
correspond to inflectional affixes (i.e. endings and markers) and
clitics. Unfortunately the parser does not distinguish derivational
affixes. The first 100~000 word tokens were used as training data, and
the following 100~000 word tokens were used as test data. The test
data contained 34~821 word types.

The English corpus consisted of mainly newspaper text from the Brown
corpus\footnote{The Brown corpus is available at the Linguistic Data Consortium at 
http://www.ldc.upenn.edu/}. A morphological analysis of the words
was performed using the Lingsoft ENGTWOL analyzer\footnote{http://www.lingsoft.fi/}.
In case of multiple alternative morphological analyses, the shortest
analysis was selected. All characters were converted to lower case,
and words containing other characters than \emph{a} through \emph{z},
an apostrophe or a hyphen were removed. Other than
morphemic tags were removed from the morphological analyses
of the words. The remaining tags correspond to inflectional or
derivational affixes. A set of 100~000 word tokens from the corpus sections
\emph{Press Reportage} and \emph{Press Editorial} were used as
training data. A separate set of 100~000 word tokens from the sections
\emph{Press Editorial}, \emph{Press Reviews}, \emph{Religion}, and
\emph{Skills Hobbies} were used as test data. The test data contained
12~053 word types.

\begin{table*}[!ht]
\caption{Test results for the Finnish and English corpus. Method names
are abbreviated: Recursive segmentation and MDL cost (Rec.~MDL), Sequential
segmentation and ML cost (Seq.~ML),
and Linguistica (Ling.).
The total MDL cost measures the compression of the corpus.
However, the cost is computed according to Equation~(\ref{eqn:MDLcost}), which
favors the Recursive MDL method.
The final number of morphs in the codebook
(\#morphs in codebook) is a measure of the size of the morph
``vocabulary''. The relative codebook cost gives the
share of the total MDL cost that goes into coding the codebook.
The alignment distance is the total distance computed over the
sequence of morph/morphemic label pairs in the test data.
The unseen aligned pairs is the percentage of all aligned morph/label pairs
in the test set that were never observed in the training set. This
gives an indication of the generalization capacity of the method to new word
forms. 
}
\label{tab:TestResults}
\vskip 0.12in
\centerline{
\begin{tabular}{|l||r|r|r||r|r|r|}
\hline
Language & \multicolumn{3}{|c||}{Finnish} &
\multicolumn{3}{c|}{English} \\
\hline
Method & Rec. MDL & Seq. ML & Ling. & Rec. MDL & Seq. ML &
 Ling. \\
\hline
\hline
Total MDL cost [bits] & \textbf{2.09M} & 2.27M & 2.88M & \textbf{1.26M} & 1.34M & 1.44M \\
\hline
\#morphs in codebook & \textbf{6302} & 10 977 & 22 075 & \textbf{3836} & 4888 & 8153 \\
\hline
Relative codebook cost & \textbf{10.16\%} & 15.27\% & 36.81\% & \textbf{9.42\%} & 10.90\% & 19.14\% \\
\hline
Alignment distance & \textbf{768k} & 817k & 1111k & \textbf{313k} & 444k & 332k \\
\hline
Unseen aligned pairs & 23.64\% & \textbf{20.20\%} & 37.22\% & \textbf{18.75\%} & 19.67\% & 20.94\% \\
\hline
Time [sec] & 620 & 390 & \textbf{180} & 130 & 80 & \textbf{30} \\
\hline
\end{tabular}}
\end{table*}

Test results for the three methods and the two languages are shown in
Table~\ref{tab:TestResults}. We observe different tendencies for
Finnish and English. For Finnish, there is a correlation between the
compression of the corpus and the linguistic generalization capacity
to new word forms. The Recursive splitting with the MDL cost function is 
clearly superior to the Sequential splitting with ML cost, which 
in turn is superior to \emph{Linguistica}. The Recursive MDL method is 
best in terms of data compression: it produces the smallest
morph lexicon (codebook), and the codebook naturally occupies a small
part of the total cost. It is best also in terms of the linguistic measure,
the total alignment distance on test data.
\emph{Linguistica}, on the other hand, employs a more
restricted segmentation, which leads to a larger codebook and to the
fact that the codebook occupies a large part of the total MDL
cost. This also appears to lead to a poor generalization ability to new word
forms. The linguistic alignment distance is the highest, and so is the percentage
of aligned morph/morphemic label pairs that were never observed in the
training set. On the other hand, \emph{Linguistica} is the 
fastest program\footnote{Note, however, that the computing time 
comparison with \emph{Linguistica} is only approximate since it was 
a compiled program run on Windows whereas the two other methods 
were implemented as Perl scripts run on Linux.}.

Also for English, the Recursive MDL method achieves the best
alignment, but here \emph{Linguistica} achieves nearly the same
result. The rate of compression follows the same pattern as for
Finnish, in that \emph{Linguistica} produces a much larger morph
lexicon than the methods presented in this paper. In spite of this
fact, the percentage of unseen morph/morphemic label pairs is about
the same for all three methods. This suggests that in a
morphologically poor language such as English a restrictive
segmentation method, such as \emph{Linguistica}, can compensate for
new word forms -- that it does not recognize at all -- with old,
familiar words, that it ``gets just right''. In contrast, the methods
presented in this paper produce a morph lexicon that is smaller and
able to generalize better to new word forms but has somewhat lower
accuracy for already observed word forms.

\paragraph{Visual inspection of a sample of words.}
In an attempt to analyze the segmentations more thoroughly, we
randomly picked 1000 different words from the Finnish test set. The
total number of occurrences of these words constitute about 2.5\% of
the whole set. We inspected the segmentation of each word visually and
classified it into one of three categories: (1) correct and complete
segmentation (i.e., all relevant morpheme boundaries were identified),
(2) correct but incomplete segmentation (i.e.,
not all relevant morpheme boundaries were identified, but no
proposed boundary was incorrect),
(3) incorrect segmentation (i.e., some proposed boundary
did not correspond to an actual morpheme boundary).

The results of the inspection for each of the three segmentation
methods are shown in Table~\ref{tab:VisualInspection}. The Recursive
MDL method performs best and segments about half of the words
correctly. The Sequential ML method comes second and
\emph{Linguistica} third with a share of 43\% correctly segmented
words. When considering the incomplete and incorrect segmentations the
methods behave differently. The Recursive MDL method leaves very
common word forms unsplit, and often produces excessive splitting for
rare words. The Sequential ML method is more prone to excessive
splitting, even for words that are not rare. \emph{Linguistica}, on the
other hand, employs a more conservative splitting strategy, but makes
incorrect segmentations for many common word forms.

\begin{table}
\caption{Estimate of accuracy of morpheme boundary detection based on
visual inspection of a sample of 2500 Finnish word tokens.}
\label{tab:VisualInspection}
\vskip 0.12in
\begin{tabular}{|l|r|r|r|}
\hline
Method & Correct & Incomplete & Incorrect \\
\hline
\hline
Rec. MDL & 49.6\% & 29.7\% & 20.6\% \\
\hline
Seq. ML & 47.3\% & 15.3\% & 37.4\% \\
\hline
Linguistica & 43.1\% & 24.1\% & 32.8\% \\
\hline
\end{tabular}
\end{table}

The behaviour of the methods is illustrated by
example segmentations in Table~\ref{tab:RealSegments}.
Often the Recursive MDL method produces complete and correct
segmentations. However, both it and the Sequential ML method can
produce excessive splitting, as is shown for the latter,
e.g. \texttt{affecti + on + at + e}.  In contrast, \emph{Linguistica}
refrains from splitting words when they should be split, e.g.,
the Finnish compound words in the table.

\begin{table*}
\caption{Some English and Finnish word segmentations produced by the
three methods. The Finnish words are
\texttt{el\"ainl\"a\"ak\"ari} (\emph{veterinarian}, lit. \emph{animal
doctor}), \texttt{el\"ainmuseo} (\emph{zoological museum},
lit. \emph{animal museum}), \texttt{el\"ainpuisto} (\emph{zoological park},
lit. \emph{animal park}), and \texttt{el\"aintarha} (\emph{zoo},
lit. \emph{animal garden}). The suffixes \texttt{-lle},
\texttt{-n}, \texttt{-on}, and \texttt{-sta} are linguistically correct. 
(Note that
in the Sequential ML method the rejection criteria mentioned are not
applied on the last round of Viterbi segmentation. This is why two one
letter morphs appear in a sequence in the 
segmentation \texttt{el\"ain + tarh + a + n}.)}
\label{tab:RealSegments}
\vskip 0.12in
\centerline{
\begin{tabular}{|l|l|l|}
\hline
Recursive MDL & Sequential ML & Linguistica \\
\hline
\hline
affect & affect & affect \\
affect + ing & affect + ing & affect + ing \\
affect + ing + ly & affect + ing + ly & affect + ing + ly \\
affect + ion & affecti + on & affect + ion \\
affect + ion + ate & affecti + on + at + e & affect + ion + ate \\
affect + ion + s & affecti + on + s & affect + ion + s \\
affect + s & affect + s & affect + s \\
\hline
\hline
el\"ain + l\"a\"ak\"ari & el\"ain + l\"a\"ak\"ari &
el\"ainl\"a\"ak\"ari \\
el\"ain + l\"a\"ak\"ari + lle &
el\"ain + l\"a\"ak\"ari + lle &
el\"ainl\"a\"ak\"ari + lle \\
el\"ain + museo + n & el\"ain + museo + n & el\"ainmuseo + n \\
el\"ain + museo + on & el\"ain + museo + on & el\"ainmuseo + on \\
el\"ain + puisto + n & el\"ain + puisto + n & el\"ainpuisto + n \\
el\"ain + puisto + sta & el\"ain + puisto + sta & el\"ainpuisto + sta \\
el\"ain + tar + han & el\"ain + tarh + a + n & el\"aintarh + an \\
\hline
\end{tabular}}
\end{table*}

\section{Discussion of the Model}

Regarding the model, there is always room for improvement. In
particular, the current model does not allow representation of
contextual dependencies, i.e., that some morphs appear only in
particular contexts (allomorphy). Moreover, languages have rules 
regarding the ordering of stems and affixes (morphotax).
However, the current model has no way of representing such contextual
dependencies.

\section{Conclusions}
\label{sec:conclusion}

In the experiments the online method with the MDL cost function and
recursive splitting appeared most successful especially for Finnish,
whereas for English the compared methods were rather equal in
performance. This is likely to be partially due to the model structure
of the presented methods which is especially suitable for languages
such as Finnish. However, there is still room for considerable
improvement in the model structure, especially regarding the
representation of contextual dependencies.

Considering the two examined model optimization methods, the Recursive
MDL method performed consistently somewhat better. Whether this is due
to the cost function or the splitting strategy cannot be deduced based
on these experiments.  In the future, we intend to extend the latter
method to utilize an MDL-like cost function.


\bibliographystyle{acl}
\bibliography{acl02}

\end{document}